# Practically Perfect

**Christopher Meek**
Microsoft Research
Redmond, WA 98052-6399
meek@microsoft.com

**David Maxwell Chickering**
Microsoft Research
Redmond, WA 98052-6399
dmax@microsoft.com

## Abstract

We prove that perfect distributions exist when using a finite number of bits to represent the parameters of a Bayesian network. In addition, we provide an upper bound on the probability of sampling a non-perfect distribution when using a fixed number of bits for the parameters and that the upper bound approaches zero exponentially fast as one increases the number of bits. We also provide an upper bound on the number of bits needed to guarantee that a distribution sampled from a uniform Dirichlet distribution is perfect with probability greater than 1/2.

## 1 INTRODUCTION

We consider the following sampling problem: if the parameters of a Bayesian network are chosen randomly, how likely is it that the resulting distribution is perfect with respect to that model? Meek (1995) and Spirtes, Glymour and Scheines (2000) consider this problem and demonstrate that in almost all cases, the sampled distribution is perfect with respect to the model. These researchers, however, do not account for the fact that, in practice, distributions are stored using a finite number of bits on a computer; instead, their results rely on using real numbers to represent the parameter values in a Bayesian network and consequently we cannot apply their results in practice.

In this paper, we extend the results of Meek (1995) and Spirtes et al. (2000) to the case where the parameters of the Bayesian network are represented using a finite number of bits. We provide an upper bound on the probability of sampling a non-perfect distribution when using a fixed number of bits for the parameters, and we show that this bound approaches zero exponentially fast as we increase the number of bits. Our results guarantee that perfect distributions with fixed-length representations exist.

The property of perfectness plays an important role in the theory of Bayesian networks. First, the existence of perfect distributions for arbitrary sets of variables and directed acyclic graphs implies that various methods for reading independence from the structure of the graph (e.g., Pearl, 1988; Lauritzen, Dawid, Larsen & Leimer, 1990) are complete. Second, the asymptotic reliability of various search methods is guaranteed under the assumption that the generating distribution is perfect (e.g., Spirtes, Glymour & Scheines, 2000; Chickering & Meek, 2002).

In addition to its theoretical importance, perfectness plays an important role in the experimental evaluation of structure-learning algorithms. In a common evaluation approach, researchers generate synthetic data from one or more Bayesian networks with known structure and sampled parameters, and then evaluate competing algorithms based on whether they can recover the generative structure more reliably and/or with fewer data records. If the generative distribution is not perfect with respect to the generative structure, however, this evaluation approach may erroneously penalize algorithms for not identifying dependencies that do not, in fact, exist in the generative distribution. Our results show that by using sufficiently many bits to represent the sampled parameter values, researchers are assured that, with high probability, the evaluation approach is appropriate.

## 2 BAYESIAN NETWORKS AND PERFECT DISTRIBUTIONS

In this paper, we concentrate on Bayesian networks for a set of variables $\mathbf{X} = \{X_1, \ldots, X_n\}$, where each $X_i \in \mathbf{X}$ has a finite number of states $r_i$. A *parametric Bayesian-network model* $\mathcal{B}$ for a set of variables $\mathbf{X} = \{X_1, \ldots, X_n\}$ is a pair $(\mathcal{G}, \beta_\mathcal{G})$. $\mathcal{G} = (\mathbf{V}, \mathbf{E})$ is a directed acyclic graph—or *DAG* for short—consisting of (1) vertices **V** in one-to-one correspondence with the variables **X**, and (2) directed edges **E** that connect the vertices. $\beta_{\mathcal{G}}$ is a set of parameter values that specify all of the conditional probability distributions. We use $\beta_{\mathcal{G}}$ to denote the parameters of a $\beta$-*parameterization* of the Bayesian network that we now describe. We use $\beta_i \subset \beta_{\mathcal{G}}$ to denote the subset of these parameter values that define the (full) conditional probability table of vertex $X_i$ given its parents in $\mathcal{G}$. $\beta_{ijk}$ ($1 \leq k \leq r_i$) denotes the conditional probability that variable $X_i = k$ (i.e., the $i^{th}$ variable is in state $k$) given (1) its parents are in the $j^{th}$ configuration and (2) $X_i \neq l$ for all $l < k$. We use $\beta_{ij} = \cup_k \beta_{ijk}$ to denote the $r_i$ conditional probabilities associated with parent configuration $j$ of variable $X_i$. Because the parameters $\beta_{ij}$ define a conditional table, there are only $r_i - 1$ independent parameters, and, by definition, $\beta_{ijr_i} = 1$.

A parametric Bayesian network represents a joint distribution over **X** that factors according to the structure $\mathcal{G}$ as follows:

$$P_{\mathcal{B}}(X_1 = x_1, \ldots, X_n = x_n) = \prod_{i=1}^{n} P(X_i = x_i | \mathbf{Pa}_i^{\mathcal{G}} = \mathbf{pa}_i^{\mathcal{G}}, \beta_i) \quad (1)$$

where $\mathbf{Pa}_i^{\mathcal{G}}$ is the set of parents of vertex $X_i$ in $\mathcal{G}$. A *Bayesian-network model* (or DAG model) $\mathcal{G}$ is simply a directed acyclic graph and represents a family of distributions that satisfy the independence constraints that must hold in any distribution that can be represented by a parametric Bayesian network with that structure. We say that a Bayesian network $\mathcal{G}$ can *represent* a distribution $P(\mathbf{X})$ if the distribution is defined by some parametric Bayesian network with structure $\mathcal{G}$.

We also use the standard parameterization given in terms of $\theta_{ijk}$ which denotes the probability that the $i^{th}$ variable is in state $k$ when its parents are in the $j^{th}$ configuration. We use $\boldsymbol{\theta}_{ij} = \cup_k \theta_{ijk}$ to denote the $r_i$ probabilities associated with parent configuration $j$ of variable $X_i$. We choose to parameterize Bayesian networks using the $\beta$-parameterization $\beta_{\mathcal{G}}$ because the parameters in this representation are variationally independent; that is, the set of possible values of each parameter is not constrained given arbitrary values of the remaining parameters. The parameters in the standard representation are not variationally independent because of the sum-to-one constraint for each of the sets of parameters $\boldsymbol{\theta}_{ij}$.

The relationship between our $\beta$-parameterization and the standard parameterization is as follows:

$$\theta_{ijk} = (1 - \sum_{l=1}^{k-1} \theta_{ijl}) \beta_{ijk}. \quad (2)$$

## 2.1 PERFECTNESS

The graphical structure of a parametric Bayesian network $\mathcal{B}$ can be used to determine independencies that must hold in the distribution represented by $\mathcal{B}$. More specifically, one can use d-separation (Pearl, 1988) to determine the set of all independence constraints imposed by the structure $\mathcal{G}$ via Equation 1. For disjoint sets **A**, **B**, **C**, we use $\text{dsep}_{\mathcal{G}}(\mathbf{A}, \mathbf{B}|\mathbf{C})$ to denote the fact that **A** is d-separated from **B** given **C** in graph $\mathcal{G}$ and $\mathbf{A} \perp\!\!\!\perp_P \mathbf{B}|\mathbf{C}$ to denote the fact that **A** is independent of **B** given **C** in distribution $P$.

**Theorem 1 (Pearl, 1988)** *If Bayesian network $\mathcal{G}$ can represent distribution $P$ then $\text{dsep}_{\mathcal{G}}(\mathbf{A}, \mathbf{B}|\mathbf{C})$ implies $\mathbf{A} \perp\!\!\!\perp_P \mathbf{B}|\mathbf{C}$.*

A distribution $P$ is perfect with respect to graph $G$ if the set of d-separation facts and independence facts coincide. More formally, a distribution $P$ is *perfect* with respect to graph $G$ if and only if for all disjoint sets **A**, **B**, **C** it is the case that $\mathbf{A} \perp\!\!\!\perp_P \mathbf{B}|\mathbf{C}$ if and only if $\text{dsep}_{\mathcal{G}}(\mathbf{A}, \mathbf{B}|\mathbf{C})$,

The following proposition is a trivial consequence of Theorem 1 and the definition of perfectness and is useful for analyzing whether or not a sampled parametric Bayesian-network model is perfect.

**Proposition 1** *Let $P$ be representable by a Bayesian network $\mathcal{G}$. $P$ is perfect if and only if $\neg\text{dsep}_{\mathcal{G}}(\mathbf{A}, \mathbf{B}|\mathbf{C})$ implies $\neg\mathbf{A} \perp\!\!\!\perp_P \mathbf{B}|\mathbf{C}$*

If $\neg\text{dsep}_{\mathcal{G}}(\mathbf{A}, \mathbf{B}|\mathbf{C})$ we say that **A** is d-connected to **B** given **C** and if $\neg\mathbf{A} \perp\!\!\!\perp_P \mathbf{B}|\mathbf{C}$ we say that **A** is dependent on **B** given **C**. Thus, to test for perfectness, we need only check that for every d-connection fact that holds in $\mathcal{G}$ there is a corresponding dependence fact that holds in $P$.

## 3 SAMPLING DISTRIBUTIONS

We concentrate on approaches to sampling the parameters of a $\beta$-parameterized Bayesian network in which the parameters $\beta_{ijk}$ are sampled independently according to some distribution. Limiting attention to $\beta$-parameterized sampling schemes, especially those in which the samples are independent, might seem restrictive as most sampling schemes for Bayesian network parameters use the standard parameterization $\boldsymbol{\theta}_{\mathcal{G}}$ and sample the parameters $\boldsymbol{\theta}_{ij} \sim \text{Dir}(\alpha_1, \ldots, \alpha_{r_i})$. As we show in Proposition 3, however, every Dirichlet sampling scheme for the standard parameterization has a corresponding independent $\beta$-parameterized sampling scheme.

When parameters are real valued, each parameter $\beta_{ijk}$

lies in the closed unit interval $[0,1]$, $\beta_{ij}$ lies in the closed $r_i$-dimensional cube $[0,1]^{r_i}$, and samples are drawn according to a density function. When using a computer to sample distributions, however, we use a finite representation for the parameters and sample the parameter values with a distribution function with finite support. We assume a $b$-bit representation in which the real values corresponding to the $b$-bit strings are uniformly spaced over the unit interval. In particular, for a $b$-bit representation, we represent the rational numbers $0/(2^B - 1), 1/(2^B - 1), \ldots, 1$ and use $b_l = l/(2^B - 1)$ ($l = 0, 1, \ldots, 2^B - 1$) to denote the possible values of a $b$-bit parameter.

In practice, sampling schemes are often defined in terms of density functions over real numbers but realized in terms of a $b$-bit sampling scheme. We provide an idealized mapping from a real-valued sampling scheme to a *derived* $b$-bit sampling scheme as follows: for a real-valued sampling scheme where $\beta_{ijk}$ is sampled according to density $f(\beta_{ijk})$, the derived $b$-bit sampling scheme is defined in terms of $P_{ijk}(\beta_{ijk} = b_l) = f(b_l)/\sum_l f(b_l)$ (for $k < r_i$). Alternatively, one could derive a $b$-bit sampling distribution in terms of definite integrals of the density for non-overlapping, exhaustive ranges of real values associated with each $b$-bit value $b_l$. Note that our method for deriving a sampling scheme roughly corresponds to using uniform-sized ranges and approximating the integral of the density over a range with the product of the width of the range and the height of the density at some point in the range. In practice, the mapping from a real-valued sampling scheme to its actual $b$-bit sampling scheme is a function of the particular implementation.

Finally, a sampling distribution for a Bayesian network is called a *$b$-bit $u$-bounded independent sampling scheme* if and only if (i) the parameters are $b$ bit numbers where $b > 1$, (ii) the parameters $\beta_{ijk}$ are sampled independently according to $P_{ijk}$ and (iii) for all $i, j, k, l$ $P_{ijk}(\beta_{ijk} = b_l) \leq u/2^b$.

We begin by showing that bounded continuous sampling densities lead to derived bounded sampling schemes.

**Proposition 2** *If the sampling density $f(\beta_{ijk})$ is continuous and bounded there exists an $m$ such that for all $b > m$, the $b$-bit derived sampling distribution is a $b$-bit $u$-bounded sampling scheme for some $u$.*

**Proof**: Let $x_{mode}$ be the mode of $f(\cdot)$. Without loss of generality we assume that $x_{mode} > 0$. From the continuity of $f(\cdot)$, we know that for some $\epsilon > 0$ and for all $x$ where $(x_{mode} - \epsilon \leq x < x_{mode})$ it is the case $f(x) \geq f(x_{mode} - \epsilon) > 0$. Under the assumption that our $b$-bit representation is uniform on the unit interval, we know that there are at least $\lfloor \epsilon(2^b - 1) \rfloor$ points ($b_i$'s) between $x_{mode} - \epsilon$ and $x_{mode}$. Also, for $b > 1$, $\lfloor \epsilon(2^b - 1) \rfloor > \epsilon 2^{b-1}$. If $\epsilon 2^{b-1} > 1$ then there is some $b_i$ such that $f(b_i) \geq f(x_{mode} - \epsilon)$.

Therefore, we know that

$$\frac{f(b_i)}{\sum_l f(b_l)} \leq \frac{f(x_{mode})}{\epsilon 2^{b-1} f(x_{mode} - \epsilon)}$$

and, thus, for large enough $b$, the distribution is $u$-bounded for $u = 2f(x_{mode})/(\epsilon f(x_{mode} - \epsilon))$. □

The following proposition is proved in Bernardo and Smith (1994; page 135).

**Proposition 3** $P(\theta_{ij}) \sim \text{Dir}(\alpha_1, \ldots, \alpha_{r_i})$ *if and only if* $P(\beta_{ij}) = \prod_{k < r_i} \text{Beta}(\alpha_k, \sum_{l > k} \alpha_l)$.

As noted above, researchers often draw samples of the standard parameters $\theta_{ij}$ from a Dirichlet distribution. From Proposition 3, we can implement such a sampling scheme using the $\beta$-parameterization by sampling each $\beta_{ijk}$ parameter independently from the appropriate Beta distribution. We call this approach the *derived Dirichlet sampling scheme*. We emphasize that this scheme does *not* sample the $\theta_{ij}$ parameters directly, but rather samples the $\beta_{ij}$ parameters in such a way that $\theta_{ij}$ is distributed as a Dirichlet. The distinction is important because our bounding results are based on the number of bits used to represent each $\beta_{ijk}$ parameter.

From Proposition 2 and Proposition 3, the derived Dirichlet sampling scheme is a $b$-bit $u$-bounded sampling scheme. We now provide a more specific version of Proposition 2 for the individual Beta samples that make up the Dirichlet sampling scheme:

**Proposition 4** *The $b$-bit sampling scheme derived from a $\text{Beta}(1, k - 1)$ distribution is a $2^{k-1}$-bounded sampling scheme for $b > 1$.*

**Proof**: Let $x \sim \text{Beta}(1, k - 1) = f(x)$. The mode of the distribution is at 0, $f(0) = k$, and $f(1/2) = k(1/2)^{k-2}$. From properties of the Beta distribution, we know that $f(x) > f(1/2)$ for $(x < 1/2)$. Under the assumption about our representation, half of the $2^b$ points ($b_i$'s) have values ($f(b_i)$'s) greater than or equal to $1/2$. Therefore, we know that

$$\frac{f(b_i)}{\sum_l f(b_l)} \leq \frac{f(0)}{2^{b-1} f(1/2)} = \frac{2^{k-1}}{2^b}$$

and, thus, for $b > 1$, the distribution is $2^{k-1}$-bounded. □

When the Dirichlet of interest is *uniform*—that is, $f(\theta_{ij}) \propto 1$—we call the corresponding sampling scheme the *derived uniform Dirichlet sampling* *scheme*. Propositions 3 and 4 lead immediately to the following result.

**Proposition 5** *A b-bit derived uniform Dirichlet sampling scheme for a Bayesian network in which each node has at most $r_{max}$ states is a b-bit $2^{r_{max}-1}$-bounded sampling scheme.*

## 4 SAMPLING PERFECT DISTRIBUTIONS

In this section, we present upper bounds on the probability of sampling a non-perfect distribution and an upper bound on the number of bits required to guarantee that a derived uniform Dirichlet sampling distribution will yield a perfect distribution with probability greater than or equal to 1/2. Proofs of theorems and corollaries are presented in Section 6.

**Theorem 2** *The probability that a distribution is not perfect when sampling from a b-bit u-bounded independent sampling scheme for a parametric Bayesian network with m variables is less than*

$$\frac{u r_{max} m^3 2^{m-1}}{2^b}.$$

Note that the numerator in Theorem 2 is constant for a fixed Bayesian network. Therefore, the theorem demonstrates that when sampling distributions for Bayesian networks, the probability of sampling a non-perfect distribution approaches 0 exponentially fast in the number of bits used to represent the parameters.

Finally, we provide an upper bound for the number of bits needed to guarantee that, when using a derived uniform Dirichlet sampling scheme, we sample a perfect distribution with at least probability 1/2.

**Theorem 3** *Let $\mathcal{G}$ be a Bayesian network containing m variables each with at most $r_{max}$ states. If the parameters in $\beta_\mathcal{G}$ are sampled from a derived uniform Dirichlet distribution, and if each parameter in $\beta_\mathcal{G}$ is represented by more than $m + 3\log_2 m + \log_2 r_{max} + r_{max} - 1$ bits, then the probability that the resulting distribution is perfect with respect to $\mathcal{G}$ is greater than $\frac{1}{2}$.*

## 5 DISCUSSION

While our results do provide an upper bound for the probability of sampling non-perfect distribution these bounds are not tight. It might be useful to provide tighter upper bounds on the probabilities of sampling non-perfect distributions and to investigate alternative classes of conditional distributions beyond full conditional tables.

Although we concentrated on Bayesian networks with discrete variables, the techniques used in this paper can also be extended to other types of Bayesian networks. For example, we could extend our techniques to sample conditional-Gaussian distributions for continuous variables.

## Acknowledgements

We would like to thank László Lovász for directing us to the Schwartz-Zippel theorem and David Heckerman for directing us to the result connecting the Dirichlet distribution to a product of Beta distributions.

## 6 APPENDIX: PROOFS

We begin by providing a proof sketch. We construct a polynomial that we call the perfect polynomial whose variables are the parameters of the parametric Bayesian network. By construction, if the polynomial is non-zero then the distribution represented by the parametric Bayesian network is perfect. We apply a generalization of the Schwartz-Zippel theorem to bound the probability of sampling parameter values such that the perfect polynomial is zero (i.e., potentially non-perfect). The bound is a function of the degree of the polynomial and the distributions for sampling the parameter values.

### 6.1 POLYNOMIALS

The *degree* of a polynomial in several variables is defined as the largest degree of its terms (monomials) and the *degree* of a monomial is the sum of the exponents of the variables in it.

The following result is a generalization of Schwartz-Zippel Theorem commonly used in the analysis of randomized algorithms (e.g., Motwani & Raghavan, 1995).

**Theorem 4** *If $f$ is a not identically 0 polynomial in $n$ variables with degree at most $k$, and the values of the variables $\psi_i$ ($i = 1, \ldots, n$) are chosen from the set $\{0/(N-1), 1/(N-1), \ldots, (N-1)/(N-1)\}$ independently of each other according to the distribution $P_i$ such that $P_i(\psi_i = l) \leq r/N$ then*

$$Prob(f(\psi_1, \ldots, \psi_n) = 0) \leq (rk)/N$$

*where $\text{Prob}(\psi) = \prod_i P_i(\psi_i)$ is the probability distribution over the values of the polynomial.*

**Proof**: We prove the assertion by induction on $n$. The statement is true for $n = 1$ because a non-zero polynomial in one variable of degree $k$ can have at most $k$ roots and each root occurs with probability at most $r/N$. Let $n > 1$ and let us arrange $f$ according to the powers of $\psi_1$:

$$f = f_0 + f_1\psi_1 + f_2\psi_1^2 + \cdots + f_t\psi_1^t,$$

where $f_0, \ldots, f_t$ are polynomials of the variables $\psi_2, \ldots, \psi_n$, the term $f_t = f_t(\psi_2, \ldots, \psi_n)$ is not identically 0, and $t \leq k$. Now,

$$\begin{aligned}&\text{Prob}(f(\psi) = 0) \\ &= \text{Prob}(f = 0 | f_t = 0)\text{Prob}(f_t = 0) + \\ &\quad \text{Prob}(f = 0 | f_t \neq 0)\text{Prob}(f_t \neq 0) \\ &\leq \text{Prob}(f_t = 0) + \text{Prob}(f = 0 | f_t \neq 0).\end{aligned}$$

We can bound the first term by the induction hypothesis, using the fact that the degree of $f_t$ is at most $k - t$; thus the first term is at most $r(k-t)/N$. For the second term we consider $f$ as a polynomial of $\psi_1$. Because $f$ is not identically 0 this polynomial when considered a polynomial of $\psi_1$ is not identically 0. In addition, because $\psi_1$ can be chosen independently of the variables $\psi_2, \ldots, \psi_n$ (i.e., no matter how the latter are fixed such that $f_t \neq 0$) the second term is at most $(rt)/N$. Hence $\text{Prob}(f(\psi_1, \ldots, \psi_n) = 0) \leq r((k-t)/N + t/N) = rk/N$. □

### 6.2 MAIN RESULTS

In this section, we construct the perfect polynomial, investigate several properties of this polynomial and provide proofs of our main results. We begin with some additional definitions.

When both $\mathbf{A}$ and $\mathbf{B}$ each contain a single variable we call $\text{dsep}(\mathbf{A}, \mathbf{B}|\mathbf{C})$ a singleton d-separation fact and $\mathbf{A} \perp\!\!\!\perp \mathbf{B}|\mathbf{C}$ a singleton independence fact. We call the negation of a singleton d-separation (independence) fact a singleton d-connection (dependence) fact. We will sometimes express singleton statements in terms of the single variable rather than using set notation. For example, $X_i \perp\!\!\!\perp_P X_j | \mathbf{C}$ is the singleton independence fact stating that $X_i$ is independent of $X_j$ given $\mathbf{C}$.

**Remark 5** *The singleton independence statement $X_i \perp\!\!\!\perp X_j | \mathbf{C}$ holds in a distribution $P$ only if the following polynomial is zero*

$$P(X_i = 0, X_j = 0, \mathbf{C} = 0)P(X_i \neq 0, X_j \neq 0, \mathbf{C} = 0)$$
$$-P(X_i \neq 0, X_j = 0, \mathbf{C} = 0)P(X_i = 0, X_j \neq 0, \mathbf{C} = 0)$$

*where $\mathbf{C} = 0$ indicates that all of the variables in $\mathbf{C}$ are set to be zero. We call this polynomial the* base polynomial *for the independence fact.*

Next we consider the degree of a base polynomial. Recall that $r_{max} = max_i\{r_i\}$ is the number of states in the variable with the most states.

**Proposition 6** *The degree of a base polynomial in terms of $\beta_\mathcal{G}$ is at most $2mr_{max}$*

**Proof**: Each of the probabilities in a base polynomial can be written as the sum of probabilities over $\mathbf{X}$. In terms of the standard parameters $\theta_\mathcal{G}$, each of these probabilities can be written as a product of $m$ parameters. From Equation 2, we see that each of the parameters in $\theta_\mathcal{G}$ is at most an $r_{max} - 1$ degree polynomial in the parameters in $\beta_\mathcal{G}$. The product of two probabilities over $\mathbf{X}$ has degree at most $2m(r_{max}-1)$. □

The next remark follows from the following counting argument. There is singleton independence fact for every possible combination of three disjoint sets $\mathbf{A}, \mathbf{B}, \mathbf{C}$ where $\mathbf{A}$ and $\mathbf{B}$ are singleton. Because the sets of variables must be disjoint, the singleton sets can be chosen in $m * (m - 1)$ ways and, for each choice of singleton sets, there are $2^{m-2}$ possible non-singleton sets.

**Remark 6** *The number of singleton independence (or dependence) facts for $m$ variables is less than $m^2 2^{(m-2)}$.*

Proposition 1 provides a means of testing whether a distribution representable by a Bayesian network is perfect: verify that all d-connection statements have a corresponding dependence statement true in the distribution. The following proposition demonstrates that only singleton d-connection and singleton dependence facts need to be tested, thus reducing the number of statements needed to verify perfectness. This allows us to use only base polynomials when constructing the perfect polynomial.

**Proposition 7** *Let $P$ be a distribution represented by a Bayesian network $\mathcal{G}$. $P$ is perfect with respect to $\mathcal{G}$ if and only if singleton d-connection $\neg \text{dsep}_{\mathcal{G}}(X_i, X_j | \mathbf{C})$ implies singleton dependence $\neg X_i \perp\!\!\!\perp_P X_j | \mathbf{C}$.*

**Proof:** Given Proposition 1 we can prove this proposition by showing that the implication between singleton d-connection in $\mathcal{G}$ and the corresponding dependence fact in $P$ entails the implication between the general d-connection and dependence facts of Proposition 1. Assume that $\neg \text{dsep}_{\mathcal{G}}(\mathbf{A}, \mathbf{B} | \mathbf{C})$. From the definition of d-separation we know that there is some singleton d-connection fact $\neg \text{dsep}_{\mathcal{G}}(A_i, B_j | \mathbf{C})$ that does hold. From our assumption about singleton facts, we know that the corresponding dependence fact $\neg A_i \perp\!\!\!\perp_P B_j | \mathbf{C}$ holds. Finally, from the contrapositive form of the decomposition independence property we can show that $\neg A_i \perp\!\!\!\perp_P B_j | \mathbf{C}$ implies $\neg \mathbf{A} \perp\!\!\!\perp_P \mathbf{B} | \mathbf{C}$. Thus, the singleton d-connection and dependence statements suffice to characterize perfectness for a distribution $P$ represented by a Bayesian network $\mathcal{G}$. □

Now we construct the *perfect polynomial* by multiplying the base polynomial from each independence statement that is not implied by d-separation. From Remark 5 and Proposition 7, we know that if the perfect polynomial is not zero then the distribution is perfect.

In order to use our generalization of the Schwartz-Zippel theorem, we need to insure that the perfect polynomial is not identically zero. This can be shown using the proof technique of Meek (1995) as follows. Each of the individual base polynomials can be shown to be not identically zero (i.e., non-trivial) by construction (see Meek 1995 for details). The zero-set of a not identically zero polynomial is measure-zero. The zero-set of the perfect polynomial is the union of all of the zero-sets of the base polynomials. Because the countable union of a set of measure-zero sets is also measure-zero, there must be a measurable set in the parameter space where the perfect polynomial is non-zero and, thus, the perfect polynomial is not identically zero.

**Remark 7** *The perfect polynomial is not identically zero.*

**Theorem 2** *The probability that a distribution is not perfect when sampling from a $b$-bit $u$-bounded independent sampling scheme for a parametric Bayesian network with $m$ variables is less than*

$$\frac{u r_{max} m^3 2^{m-1}}{2^b}.$$

**Proof:** A base polynomial has a degree less than $2mr_{max}$ and the perfect polynomial is a product of fewer than $m^2 2^{m-2}$ base polynomials. Therefore, the perfect polynomial has a degree less than $r_{max} m^3 2^{m-1}$. By applying Theorem 4 using the assumption of $u$-boundedness and Remark 7 with $N = 2^b$, the result is shown. □

**Theorem 3** *Let $\mathcal{G}$ be a Bayesian network containing $m$ variables each with at most $r_{max}$ states. If the parameters in $\beta_{\mathcal{G}}$ are sampled from a derived uniform Dirichlet distribution, and if each parameter in $\beta_{\mathcal{G}}$ is represented by more than $m + 3\log_2 m + \log_2 r_{max} + r_{max} - 1$ bits, then the probability that the resulting distribution is perfect with respect to $\mathcal{G}$ is greater than $\frac{1}{2}$.*

**Proof:** We use Proposition 5 and apply Theorem 2 where $u = 2^{r_{max}-1}$ to obtain the bound that the probability of sampling a non-perfect distribution is less than $2^{r_{max}-1} r_{max} m^3 2^{m-1} / 2^b$. The result then follows from simple algebra. □